\RequirePackage{mathtools}
\documentclass[runningheads]{llncs}
\usepackage[subpreambles=true]{standalone}
\usepackage[utf8]{inputenc}
\usepackage[english]{babel}
\usepackage{algorithm}% http://ctan.org/pkg/algorithm
\usepackage[caption=false]{subfig}
\usepackage{graphicx}
\usepackage{multirow}
\usepackage{siunitx} % Required for alignment
\newcommand\trajfigheight{0.2}
\newcommand\trajfigwidth{0.4}
\newcommand\addtrajectories[3]{
\begin{figure}[!htb]
\centering
	\subfloat[Sherbrooke\label{#3:sherbrooke_#1}]{%
        \includegraphics[height = \trajfigheight\textwidth,width = \trajfigwidth\textwidth]{figures/sherbrooke_gt_#1_trajectories-crop.pdf}
    }
    \subfloat[Rouen\label{#3:rouen_#1}]{%
        \includegraphics[height = \trajfigheight\textwidth,width = \trajfigwidth\textwidth]{figures/rouen_gt_#1_trajectories-crop.pdf}
    }
    \\ \vspace{-0.8\baselineskip}
    \subfloat[St-Marc\label{#3:stmarc_#1}]{%
        \includegraphics[height = \trajfigheight\textwidth,width = \trajfigwidth\textwidth]{figures/stmarc_gt_#1_trajectories-crop.pdf}
    }
    \subfloat[René-Lévesque\label{#3:rene_#1}]{%
        \includegraphics[height = \trajfigheight\textwidth,width = \trajfigwidth\textwidth]{figures/rene_gt_#1_trajectories-crop.pdf}
    }
    \caption{#2}
    \label{#3}
\end{figure}
}

\newcommand{\MSE}{\mathrm{MSE}}
\newcommand{\mean}{\mathrm{mean}}
\newcommand{\STD}{\mathrm{STD}}

\begin{document}
\title{Road User Abnormal Trajectory Detection using a Deep Autoencoder}
% What is orcidID?
%
\author{Pankaj Raj Roy \and
Guillaume-Alexandre Bilodeau}
\authorrunning{Roy and Bilodeau}
\institute{LITIV lab., Dept. of computer \& software eng.\\Polytechnique Montréal\\
\email{\{pankaj-raj.roy,gabilodeau\}@polymtl.ca}}
\maketitle              % typeset the header of the contribution
\begin{abstract}
In this paper, we focus on the development of a method that detects abnormal trajectories of road users at traffic intersections. The main difficulty with this is the fact that there are very few abnormal data and the normal ones are insufficient for the training of any kinds of machine learning model. To tackle these problems, we proposed the solution of using a deep autoencoder network trained solely through augmented data considered as normal. By generating artificial abnormal trajectories, our method is tested on four different outdoor urban users scenes and performs better compared to some classical outlier detection methods.

\end{abstract}
%
%
%
%=======================================================================================================================
\section{Introduction}
Abnormal event detection has been an intriguing research subject for many years, namely because of the fact that the definition of an abnormal event can be very unclear. The classification of abnormal events can be a challenging task when the number of possible abnormalities can easily exceed our knowledge of abnormal behaviors. Also, it is usually a very tedious job to obtain abnormal data. There can be many different possibilities of abnormalities and some of them can even be subjective. In order to tackle this problem, many authors suggest the hypothesis that all the abnormalities are outliers of the normality distribution. By taking that into account, one can notice that abnormalities are context/scene dependent and that a uniform definition of abnormal events cannot be generalized for all kinds of scenarios other than the assumption of these being the opposite of normal events.

The goal of this paper is to detect abnormal trajectories at road intersections where the objects of interest can be identified as pedestrians, cyclists or vehicles. The problem with trajectory data at intersections is that it does not follow any particular probabilistic rules. Therefore, the best way for classifying them is to use statistical or machine learning approaches that can learn the normal trajectories in an unsupervised manner and be trained to detect outliers, which can be classified as abnormal trajectories. Various statistical approaches can solve the issue of trajectory anomaly detection. But, this problem becomes challenging when the dataset is not large enough for trajectory data classification. In this work, we want to devise a method that can use a small number of trajectory samples and that gives the best precision on the classification of normal and abnormal trajectories compared to other outliers detection methods. 

To solve the abnormal event detection problem, our contribution is to propose a deep autoencoder model coupled with a data augmentation method. This allows us to encode information about normal trajectories while removing irrelevant information. Results show that our proposed method outperforms classical methods such as one-class support vector machine (SVM).

%=======================================================================================================================
\section{Related work}
In the work of Mousavi et al. \cite{ref_article1}, the detection of abnormal crowd behavior is based on Histograms of Oriented Tracklets (HOT), the Latent Dirichlet Allocation (LDA) and an SVM that is used for the classification of frames. A spatio-temporal model based on background subtraction in the form of a co-occurrence matrix is proposed in \cite{ref_article2} for detecting objects following suspicious paths. This method cannot detect abnormal events overlapping normal paths. The Mixture of Dynamic Textures (MDT), in \cite{ref_article3}, is used for the anomaly detection in crowded scenes by modeling the appearance and the representation of dynamics of the scenes. This method is not designed to interpret trajectories. In \cite{ref_article4}, an interaction energy potentials method is proposed. An SVM is used in order to classify the abnormal activities based on the standard bag-of-word representation of each video clip. In \cite{ref_article5}, detection is implemented using a dictionary of normal patterns represented by the training features from a region in a video, which are the 3D gradient features of a spatial-temporal cubes of five frames. This method cannot be used to interpret trajectories. The covariance matrix for optical flow based feature is applied in \cite{ref_article6}. In \cite{ref_article7}, the authors demonstrated how a Convolutional Neural Network (CNN), a Long Short Term Memory (LSTM) and an SVM can be learned from very few videos and used for detecting abnormal events. Even though this method achieves 95 \% accuracy, the abnormal activity detection is only limited to binary classification and is not applicable to abnormal trajectory. The convolutional auto-encoder (CAE), proposed in \cite{ref_article8}, learns the signature of normal events with the training data. This method takes the frames, the appearance features extracted via Canny edge detector and the motion features from optical flow in the input of the CAE model and outputs the regularization of reconstruction errors (RRE) for each frame.

Abnormal event detection is fundamentally an outlier detection problem. Therefore, any classical outlier detection approach can be used. In \cite{ref_Scholkopf}, a One-Class SVM algorithm is proposed for unlabeled data, which can be used for the novelty detection. This latter form of detection is basically an outlier detector that does not require any abnormal data during the training process of the classification model \cite{PIMENTEL2014215}. In \cite{ref_isolationforest}, the Isolation Forest method isolates anomalies instead of profiling normal points and also works well with large datasets.

%=======================================================================================================================
\section{Proposed Method}
\begin{figure}[!htb]
\centering
\includegraphics[width=1\textwidth]{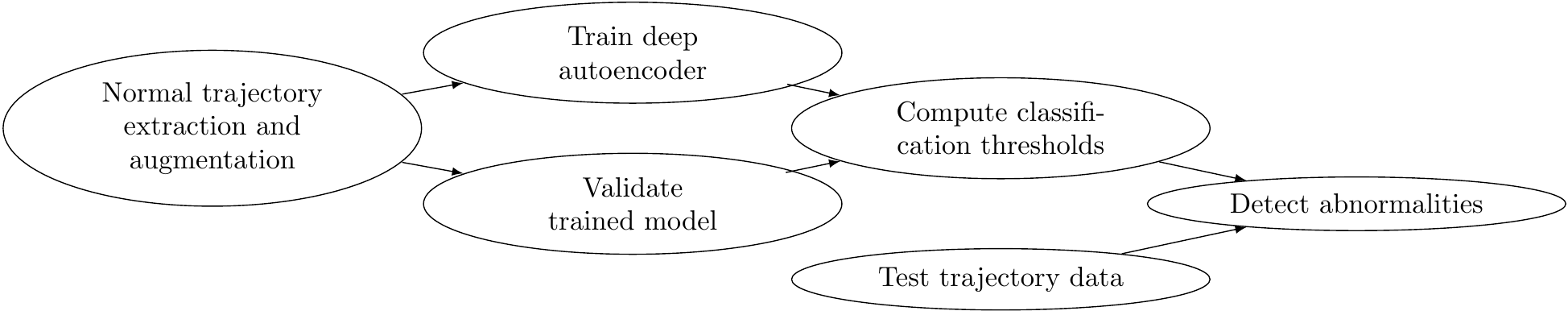}
\caption{The main steps of our abnormal trajectory detection method.} \label{fig1}
\end{figure}

Before applying any statistical or machine learning methods for the detection of outliers, the input data is processed in order to extract its trajectories. Unlike the standard classification problem where each frame of a video represents a sample, in the case of trajectory, multiple frames are needed to define a trajectory sample. Even for thousands of frames, the number of trajectory samples extracted will be insufficient for training properly any statistical or machine learning methods, especially Neural Networks (NN) based methods. Therefore, it is necessary to apply data augmentation for increasing the number of trajectory samples. The general idea behind our abnormal trajectory detection is the following (see Figure \ref{fig1}). First, tracking results are used to extract trajectories assumed to be normal. Then, data augmentation techniques are used in order to increase the number of normal trajectories. A deep autoencoder (DAE) is then trained to learn normal trajectories and classification can be applied on new data.

\subsection{Background on Autoencoders and Deep Autoencoders}
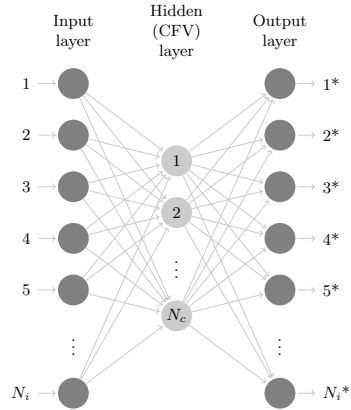
\begin{figure}[!htb]
\centering
\pagestyle{empty}
\def\layersep{2cm}
\def\nodeinlayersep{1cm}
\begin{tikzpicture}[
   shorten >=1pt,->,
   draw=black!20,
    node distance=\layersep,
    every pin edge/.style={<-,shorten <=1pt},
    neuron/.style={circle,fill=black!50,minimum size=17pt,inner sep=0pt},
    input neuron/.style={neuron},
    output neuron/.style={neuron},
    hidden neuron/.style={neuron, fill=black!20},
    annot/.style={text width=4em, text centered}
]
    % Draw the input layer nodes
    \foreach \name / \y in {1,...,7} {
      \ifnum \y=6
        \node at (0,-\y-0.5) {$\vdots$};
      \else
        \ifnum \y=7
          \node[input neuron, pin=left:$N_i$] (I-\name) at (0,-\y-0.5) {};
        \else
          \node[input neuron, pin=left:\y] (I-\name) at (0,-\y-0.5) {};
        \fi
      \fi
    }

    % Draw the hidden layer nodes
    \foreach \y in {1,...,4} {
      \ifnum \y=3
        \node at (\layersep,{-(\y+2)*\nodeinlayersep}) {$\vdots$};
      \else
        \ifnum \y=4
          \node[hidden neuron] (H1-\y) at (\layersep,{-(\y+2)*\nodeinlayersep}) {$N_c$};
        \else
          \node[hidden neuron] (H1-\y) at (\layersep,{-(\y+2)*\nodeinlayersep}) {\y};
        \fi
      \fi
    }
    % Draw the output layer node
    \foreach \y in {1,...,7} {
      \ifnum \y=6
        \node at (2*\layersep,{-(\y+1/2)*\nodeinlayersep}) {$\vdots$};
      \else
        \ifnum \y=7
          \node[output neuron, pin={[pin edge={->}]right:$N_i$*}] (O-\y) at (2*\layersep,{-(\y+1/2)*\nodeinlayersep}) {};
        \else
          \node[output neuron, pin={[pin edge={->}]right:\y*}] (O-\y) at (2*\layersep,{-(\y+1/2)*\nodeinlayersep}) {};
        \fi
      \fi
    }
    % Connect every node in the input layer with every node in the
    % hidden layer.
    \foreach \source in {1,...,5,7}
        \foreach \dest in {1,...,2,4}
            \path (I-\source) edge (H1-\dest);
    % Connect every node in the hidden layer with the output layer
    \foreach \source in {1,...,2,4}
        \foreach \dest in {1,...,5,7}
        	\path (H1-\source) edge (O-\dest);
    % Annotate the layers
    \node[annot,above of=I-1, node distance=1cm] (il1) {Input layer};
    \node[annot,right of=il1] (hl1) {Hidden (CFV) layer};
    \node[annot,right of=hl1] {Output layer};
\end{tikzpicture}
% End of code
%     without .tex extension
\caption{Diagram showing a Vanilla Autoencoder.} \label{fig2}
\end{figure}

The simplest form of autoencoder (AE), called Vanilla Autoencoder (VAE), can be used for the abnormal trajectory detection. Regular AEs are made out of fully-connected NNs that are configured to reproduce the inputs to the outputs of the network. They follow an unsupervised learning method which does not require input data to be labeled. First, this type of NNs compresses the original input into a latent-space representation, which is the space that lies in the bottleneck (hidden) layer, also known as Compressed Feature Vector (CFV). This latter form of representation is produced by an encoding function representing the encoder. Then, the decoder side of this NN reconstructs the CFV input into the output resembling the original input of the encoder side. Figure \ref{fig2} shows the simplest form of an AE, which is basically made out of three layers, one input, one hidden layer (CFV) and one reconstructed output.

The advantage of utilizing this kind of NNs is that it can be used for learning the normality of the input data. By using only normal data for training of the AE network, the encoder and the decoder are adapted for the normal data and will produce a reconstructed data where the Mean Square Error (MSE) with the original one varies within some specific defined range, which will be elaborated in the next subsection. When testing the trained AE with abnormal input data, it will produce a reconstruction error (RMSE) exceeding the threshold value calculated from the trained and validated RMSE values.

\subsubsection{Deep Autoencoder (DAE)}
Even though the VAE can be used for the detection of anomalies, intuitively, it is not enough for learning the complex nature of normal trajectories in road intersections. By adding multiple hidden layers in the AE network, the model is able to learn a more complex feature representation, which can be useful for classifying more realistic abnormal trajectories.

\subsection{Abnormal Event Detection with a Deep Autoencoder}
Algorithms \ref{alg2} and \ref{alg3} summarize the main steps required for the training of the DAE model and for detecting abnormalities using the trained network. During the training process of the DAE model, there are two types of validation samples: one that is used internally through cross-validation during the fitting of the networks with the input training set, called $va_{cv}$ samples, and the other called $va_e$ is used externally to validate the scoring of the normality of normal input data. Note that the cross-validation data is necessary during the training process of the network in order to prevent over-fitting.

\begin{algorithm}[!htb] % enter the algorithm environment
\caption{Training of the Deep Autoencoder Model.} % give the algorithm a caption
\label{alg2} % and a label for \ref{} commands later in the document
\textbf{Input:} normal data $D_n$ 
\begin{enumerate}
\item Shuffle the samples of $D_n$.
\item Split the shuffled samples of $D_n$ into training $tr$ and validation sets $va$.
\item Build the DAE model.
\item Apply normalization scaling technique to the training $tr$ and validation sets $va$ samples.
\item Fit the model with $tr$ samples by using $P_{va_{cv}}$ \% of it as the cross-validation set $va_{cv}$ in order to avoid over-fitting of the model.
\item Get the scores $S_{tr}$ of the fitted model with $tr$ samples.
\item Get the scores $S_{va}$ of the fitted model with $va$ samples.
\item Compute the threshold $\tau$ with the training and validation scores ($S_{tr}$ and $S_{va}$).
\end{enumerate}
\end{algorithm}

\begin{algorithm}[!htb] % enter the algorithm environment
\caption{Detection of Abnormal Trajectories using Trained Autoencoder.} % give the algorithm a caption
\label{alg3} % and a label for \ref{} commands later in the document
\textbf{Input:} Normal and abnormal data $D_{an}$, trained DAE model $i$ \\
\textbf{Output:} classification decisions
\begin{enumerate}
\item Apply normalization scaling method to the test trajectory samples in $D_{an}$ between 0 and 1 ($z_{scaled}$).
\item Open the saved trained model $i$ and the corresponding computed threshold value $\tau$.
\item Get the scores $S_{z}$ of the trained model with sample $z$ in $D_{an}$.
\item Detect the abnormalities using $S_{z}$ and the previously computed threshold $\tau$ and output the results.
\end{enumerate}
\end{algorithm}

The scores $S$ are based of the MSE between normalized original $z$ and reconstructed output $\hat{z}$ for each sample, as shown in equation \ref{equ1}.

\begin{equation}
S = \MSE\left( z, \hat{z} \right) \label{equ1}
\end{equation}

The threshold value $\tau$ is learned from the data and is used for separating normal from abnormal data. It is determined through equation \ref{equ2}, where STD stands for standard deviation, and, $ S_{tr} $ and $ S_{va} $ are the scores for the training and the validation sets respectively.

\begin{equation}
\tau = \mean\left( S_{tr} \right) + \mean\left( S_{va} \right) + 3 \times \left( \STD\left( S_{tr} \right) + \STD\left( S_{va} \right) \right) \label{equ2}
\end{equation}

Then, the following rule applies for the classification $C$ of a trajectory sample $z$:

\begin{equation} \label{equ3}
C(z) = 
\begin{dcases}
	\textnormal{Normal}, & \textnormal{if } S_z \leq \tau \\
    \textnormal{Abnormal}, & \textnormal{otherwise,}
\end{dcases}
\end{equation}

where $S_z$ is the score obtained by applying the trained DAE model to the trajectory sample $z$.

%=======================================================================================================================
\section{Experiments}

\subsection{Deep Autoencoder Implementation Details}
The table \ref{tab1} presents the values of hyper-parameters that resulted in the best convergence of the deep autoencoder network. The implementation of the DAE was done using Keras Python package.

\begin{table}
\centering
\caption{Hyper-parameters used for the training of the DAE.}\label{tab1}
\begin{tabular}{c|c|l}
\hline
Parameter &  Value & Definition\\
\hline \hline
$H_i$ &  125 & Input trajectory sample size \\
$H_{h_1}$ &  128 & Number of units in the first hidden layer\\
$H_{h_2}$ &  64 & Number of units in the second hidden layer\\
$H_{h_3}$ &  32 & Number of units in the third hidden layer\\
$H_{h_4}$ &  16 & Number of units in the fourth hidden layer\\
$H_c$ &  8 & Number of units in the CFV layer\\
$N_{batch}$ &  128 & Batch size used each time\\
$N_{epoch}$ &  100 & Number of epochs\\
optimiser &  RMSprop & RMSProp optimizer\\
$\sigma$ &  0.001 & Default learning rate of RMSProp optimizer\\
loss &  MSE & Mean Squared Error\\
\hline
\end{tabular}
\end{table}

Notice that the first hidden layer in the encoder side of the DAE has a number of units which is slightly greater than the input size. This configuration helps to transform the input size into a number that is a power of two. In fact, by having large layer size in the first hidden layer, the network learns a more detail representation of the input data. Then, by having decreasing unit sizes of power of two in the subsequent layers, the CFV layer gets a more abstract representations of the normal trajectory samples, which can help to detect more complex level of outliers. Also, note that the hidden layers use ReLu activation and the output layer use Sigmoid. 

Before feeding the input data into the network, it is scaled with a min and max scaler of 0 and 1. We apply a normalization scaling technique, because the input trajectory data does not follow any specific probabilistic distribution. Also, by scaling the input data before feeding it to the DAE network, we avoid the possibility of having the knock-on effect on the ability to learn the normality of the training data.

\subsection{Experimental Protocol}
We applied $N$ iterations of Repeated random sub-sampling validation for validating our model, because the shuffling of the input normal data, combined with the splitting into training and validation sets, has an impact on the accuracy of the normality/abnormality detection. Therefore, by doing $N$ iterations, we can determine the average of our method's performance to assess its power of generalization and get the best model giving the highest values of True Positive (TP) and True Negative (TN), and the lowest values of False Positive (FP) and False Negative (FN). We compared our method with OC-SVM and IF that are implemented in \texttt{sklearn} Python package \cite{scikit-learn}.

\subsection{Input Data, Data Augmentation and Processing}
We used the Urban Tracker Dataset \cite{urban_tracker} that contains four different types of scenarios involving pedestrians, vehicles (including cars and large vehicles) and bikes. Figure \ref{fig5} shows all the original trajectories of the four scenarios, all of which are considered as "normal" trajectories.

%-------------------
\addtrajectories{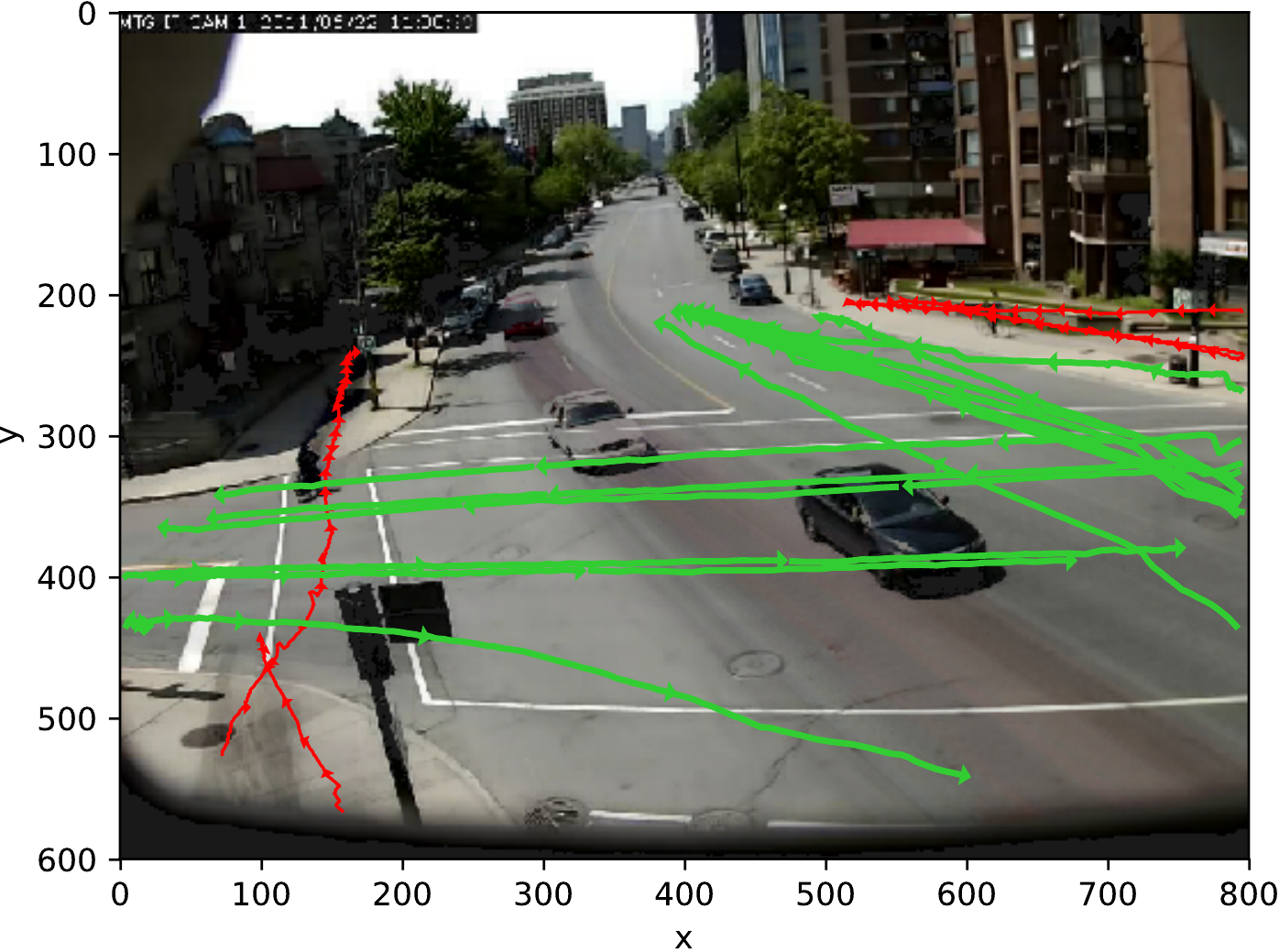}{Original annotated trajectories of the Urban Tracker Dataset. Red, green and blue represent pedestrians, cars and bikes respectively.}{fig5}
%-------------------

The Sherbrooke video contains the trajectories of 15 cars and 5 pedestrians. The Rouen video includes 4 vehicles, 1 bike and 11 pedestrians. For the St-Marc video, 1000 frames were chosen, with 7 vehicles, 2 bicycles and 19 pedestrians. Finally, the René-Lévesque video includes 29 vehicles and 2 bikes. We used the ground truth locations of the moving objects for every frames of the urban intersection videos. These locations are composed of bounding box positions. Algorithm \ref{alg1} describes the fundamental steps for the extraction and augmentation of the trajectories.

\begin{algorithm}[!htb] % enter the algorithm environment
\caption{Trajectory Extraction and Augmentation.} % give the algorithm a caption
\label{alg1} % and a label for \ref{} commands later in the document
\textbf{Input:} bounding boxes \\
\textbf{Output:} trajectory samples
\begin{enumerate}
\item Extract the positions $x$ and $y$ corresponding to the center of the bounding box positions and its related velocities $v_x$ and $v_y$ of the object $n$.
\item Generate 50 augmented trajectories by randomly generating positions $x_a$ and $y_a$ around the real ones $x$ and $y$ and its related velocities $v_{x_a}$ and $v_{y_a}$.
\item Decompose the extracted and augmented trajectories into sub trajectory frames composed of 31 positions and velocities.
\begin{itemize}
\item Stretch the complete trajectory of the object $n$ in order for it to be decomposable according to the specific sliding stride value.
\end{itemize}
\item Add the appropriate label depending on the type of the object. The label "0" identifies a pedestrian, "1" is for a car and "2" specifies a bike.
\item Flatten all the sub trajectories in the following Packed format (125 elements): $[\textnormal{label}, x_1, y_1, v_{x_1}, v_{y_1}, x_2, y_2, v_{x_2}, v_{y_2}, ..., x_{31}, y_{31}, v_{x_{31}}, v_{y_{31}}]$. 
\end{enumerate}
\end{algorithm}

Note that we choose to generate trajectory samples by applying a sliding-window approach in which the trajectory window, composed of 31 positions, slides the complete trajectory of an object with a stride of 10 frames. This is done to learn the continuity of the trajectory and therefore prevents the network to learn the wrong representation of trajectory coordinates.

\subsection{Abnormal Trajectory Generation}
For testing the trained model, abnormal trajectory data is needed. Abnormal events can be a different trajectory path for a car, or a car following the path of pedestrians. For demonstrating the validity of our method, we have generated two types of abnormal data, one which consists of straight lines with constant velocities and the other, more realistic abnormal trajectories inspired by the real normal ones. The realistic ones are resulting from the transformation of the original trajectories, which keeps the same degree of fluctuation in the positions and the velocities. The figures \ref{fig7} shows some of the generated abnormalities.

%-------------------
\addtrajectories{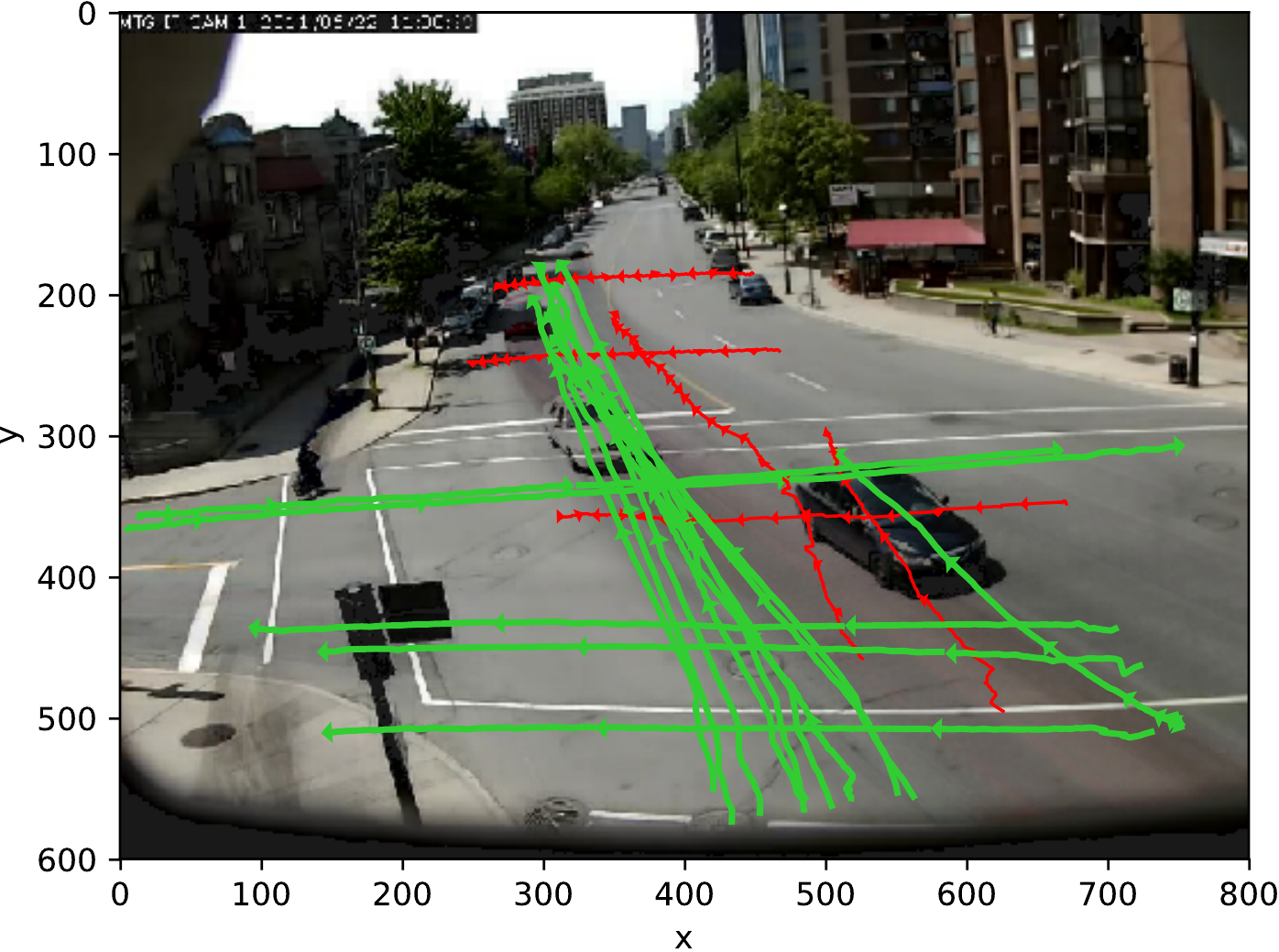}{Generated realistic abnormal trajectories. Red, green and blue represent pedestrians, cars and bikes respectively.}{fig7}
%-------------------

\subsection{Experimental Results}
Table \ref{tab2} presents the results obtained by applying the trained models on the normal and abnormal trajectories of the dataset. The true positive rate (TPR) and the false positive rate (FPR) are defined as the detected normality/abnormality from normal/abnormal and abnormal/normal samples respectively. Here, we only considered the realistic version of the generated abnormal data. Also, note that the data augmentation technique is applied before training each of these models.

\begin{table}[!htb]
\centering
\caption{Results obtained by applying the trained model on trajectory samples. Boldface values indicate the best results. Status: label indicating normal/abnormal samples, Size: number of samples, OC-SVM: One-class SVM, IF: Isolation forest, VAE: Vanilla AE, DEA: Deep AE.}
\label{tab2}
\begin{tabular}{c|c|c||c|c|c|c|c|c|c|c}
\multicolumn{3}{c}{} & \multicolumn{8}{c}{Method (\%)} \\ \cline{4-11}
\multicolumn{3}{c}{} & \multicolumn{2}{|c|}{OC-SVM} & \multicolumn{2}{c|}{IF} & \multicolumn{2}{c|}{VAE} & \multicolumn{2}{c}{DAE (ours)} \\ \hline
Data & Status & Size & \textnormal{TPR} & \textnormal{FPR} & \textnormal{TPR} & \textnormal{FPR} & \textnormal{TPR} & \textnormal{FPR} & \textnormal{TPR} & \textnormal{FPR} \\ \hline \hline
\multicolumn{1}{c|}{\multirow{2}{*}{Sherb.}} & normal & 20606 & 90 & 83 & 89 & 82 & 99 & \textbf{98} & 99 & \textbf{20} \\
\multicolumn{1}{c|}{} & abnormal & 406 & 17 & 10 & 18 & 11 & \textbf{2} & 1 & \textbf{80} & 1 \\
\hline
\multicolumn{1}{c|}{\multirow{2}{*}{Rouen}} & normal & 11884 & 90 & 87 & 90 & 91 & \textbf{100} & \textbf{96} & \textbf{100} & \textbf{15} \\
\multicolumn{1}{c|}{} & abnormal & 234 & 13 & 10 & 9 & 10 & \textbf{4} & \textbf{0} & \textbf{85} & \textbf{0} \\
\hline
\multicolumn{1}{c|}{\multirow{2}{*}{St-Marc}} & normal & 40139 & 90 & 82 & 90 & 79 & \textbf{99} & 99 & \textbf{99} & \textbf{68} \\
\multicolumn{1}{c|}{} & abnormal & 789 & 18 & 10 & 21 & 10 & 1 & \textbf{1} & \textbf{32} & \textbf{1} \\
\hline
\multicolumn{1}{c|}{\multirow{2}{*}{Rene-L.}} & normal & 45341 & 90 & 80 & 89 & 80 & \textbf{99} & 99 & \textbf{99} & \textbf{61} \\
\multicolumn{1}{c|}{} & abnormal & 891 & 20 & 10 & 20 & 10 & 1 & \textbf{1} & \textbf{39} & \textbf{1} \\
\hline
\end{tabular}
\end{table}

Globally, our method outperforms others in terms of TPR and FPR in both normal and abnormal samples. Specifically, DAE gives the smallest values of FPR compared to others. Also, the results show clearly that DAE distinguishes better the normal and abnormal data compared to VAE. Therefore, it is necessary to have multiple layers in AE network in order for it to avoid getting over-fitted by the training normal data.

%=======================================================================================================================
\section{Conclusion}
In this paper, we studied the detection of abnormal trajectories in common traffic scenarios. Considering the hypothesis of abnormalities behaving as outliers, we have proposed a method with a DAE which uses only normal data in the training process. We also applied an automated data augmentation technique for increasing the number of training samples. By generating interactively abnormal realistic trajectories, our method, compared to others like OC-SVM, IF and VAE, yielded the best performance in terms of TPR and FPR of the normal/abnormal detection in most videos.

%=======================================================================================================================
\section*{Acknowledgement}
This research was supported by a grant from IVADO funding program for fundamental research projects.

\bibliographystyle{splncs04}
\bibliography{samplepaper}
\end{document}